%% file: main.tex
\documentclass[sigconf]{acmart}

\setcopyright{acmlicensed}
\copyrightyear{2018}
\acmYear{2018}
\acmDOI{XXXXXXX.XXXXXXX}
\acmConference[Conference acronym 'XX]{Make sure to enter the correct
  conference title from your rights confirmation email}{June 03--05,
  2018}{Woodstock, NY}
\acmISBN{978-1-4503-XXXX-X/2018/06}

\AtBeginDocument{%
  \providecommand\BibTeX{{%
    \normalfont B\kern-0.5em{\scshape i\kern-0.25em b}\kern-0.8em\TeX}}}

\usepackage{colortbl}
\definecolor{lGray}{gray}{0.9}

\theoremstyle{definition}

\usepackage{caption}
\usepackage{subcaption}
\usepackage[utf8]{inputenc}
\usepackage{pgfplots}
\DeclareUnicodeCharacter{2212}{−}
\usepgfplotslibrary{groupplots,dateplot}
\usetikzlibrary{patterns,shapes.arrows}
\pgfplotsset{compat=newest}
\usepackage{hhline}
\usepackage{multirow}
\usepackage{enumitem}
\usepackage{svg}
\usepackage{adjustbox}
\usepackage{import}
\usepackage{balance}

\expandafter\def\expandafter\UrlBreaks\expandafter{\UrlBreaks
    \do\a\do\b\do\c\do\d\do\e\do\f\do\g\do\h\do\i\do\j%
    \do\k\do\l\do\m\do\n\do\o\do\p\do\q\do\r\do\s\do\t%
    \do\u\do\v\do\w\do\x\do\y\do\z\do\A\do\B\do\C\do\D%
    \do\E\do\F\do\G\do\H\do\I\do\J\do\K\do\L\do\M\do\N%
    \do\O\do\P\do\Q\do\R\do\S\do\T\do\U\do\V\do\W\do\X%
    \do\Y\do\Z\do\/\do-}

\begin{document}

\title{NILMFormer: Non-Intrusive Load Monitoring that Accounts for Non-Stationarity}

\author{Adrien Petralia}
\orcid{0000-0003-2987-9111}
\affiliation{\small
  \institution{EDF R\&D - Université Paris Cité}
  \city{Paris}
  \country{France}
}
\email{adrien.petralia@gmail.com}

\author{Philippe Charpentier}
\orcid{0000-0002-3039-2485}
\affiliation{\small
  \institution{EDF R\&D}
  \city{Palaiseau}
  \country{France}
}
\email{philippe.charpentier@edf.fr}

\author{Youssef Kadhi}
\orcid{0009-0007-1995-994X}
\affiliation{\small
  \institution{EDF R\&D}
  \city{Palaiseau}
  \country{France}
}
\email{youssef.kadhi@edf.fr}

\author{Themis Palpanas}
\orcid{0000-0002-8031-0265}
\affiliation{\small
  \institution{Université Paris Cité}
  \city{Paris}
  \country{France}
}
\email{themis@mi.parisdescartes.fr}

\renewcommand{\shortauthors}{A. Petralia et al.}

\begin{abstract}
    Millions of smart meters have been deployed worldwide, collecting the total power consumed by individual households. 
    Based on these data, electricity suppliers offer their clients energy monitoring solutions to provide feedback on the consumption of their individual appliances.
    Historically, such estimates have relied on statistical methods that use coarse-grained total monthly consumption and static customer data, such as appliance ownership.
    Non-Intrusive Load Monitoring (NILM) is the problem of disaggregating a household's collected total power consumption to retrieve the consumed power for individual appliances.
    Current state-of-the-art (SotA) solutions for NILM are based on deep-learning (DL) and operate on subsequences of an entire household consumption reading.
    However, the non-stationary nature of real-world smart meter data leads to a drift in the data distribution within each segmented window, which significantly affects model performance.
    This paper introduces NILMFormer, a Transformer-based architecture that incorporates a new subsequence stationarization/de-stationarization scheme to mitigate the distribution drift and that uses a novel positional encoding that relies only on the subsequence's timestamp information.
    Experiments with 4 real-world datasets show that NILMFormer significantly outperforms the SotA approaches. 
    Our solution has been deployed as the backbone algorithm for EDF's (Electricité De France) consumption monitoring service, delivering detailed insights to millions of customers about their individual appliances' power consumption.
    This paper appeared in KDD 2025.
\end{abstract}

\ccsdesc[500]{Hardware~Energy metering}
\ccsdesc[500]{Computing methodologies~Neural networks}

\keywords{Non-Intrusive Load Monitoring, Deep-Learning, Non-Stationarity}



\maketitle

\section{Introduction}
\label{sec:intro}
    
\begin{figure}[tb]
    \centering
    \includegraphics[width=1\linewidth]{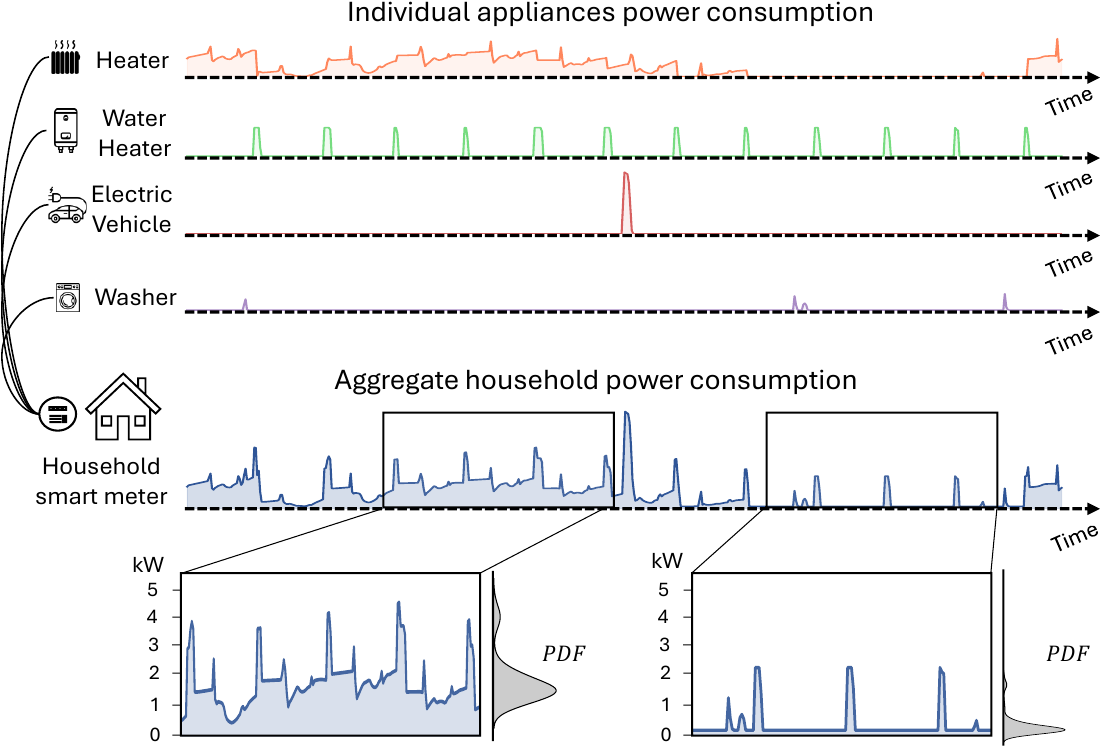}
    \caption{Illustration of a smart meter signal, composed of the addition of each individual appliance running in the household, and its non-stationary nature.} 
    \label{fig:intro} 
\end{figure}


Efficient energy management is emerging as a major lever for tackling climate change.
Thus, enabling consumers to understand and control their own energy consumption is becoming crucial. 
In this regard, millions of smart meters have been deployed worldwide in individual households~\cite{smart_meter_deployments_ue, smart_meter_deployments_us}, recording the total household electricity consumption at regular intervals (typically the average power consumed every 10-30min, depending on the country~\cite{ZHAO2020114949}), providing valuable data that suppliers use to forecast energy demand. 
Though, as Figure~\ref{fig:intro} shows, this recorded signal is the aggregated power consumed by \emph{all} appliances operating simultaneously. 

EDF (Electricité De France), one of the world's largest electricity producers and suppliers, offers to clients a consumption monitoring solution called \emph{Mon Suivi Conso}, accessible via a web interface and a mobile application~\cite{SuiviConsoEDF}. 
A key feature of this service is providing customers with details about 
individual appliance consumption.
Historically, only annual appliance-level feedback was available, and estimates relied on semi-supervised statistical methods that used the customers’ static information (e.g., appliance presence) and total monthly consumption~\cite{patentEDF1, patentEDF2}. 
However, recent user feedback has increasingly highlighted the wish to have access to 
fine-grained information~\cite{EDFprivatecomunicationdatanumia}, prompting EDF to develop a more detailed, accurate, and personalized solution for appliance consumption feedback to enhance customer satisfaction and retention.

A recent internal study at EDF explored Time Series Extrinsic Regression (TSER) approaches~\cite{tser_2021} to provide monthly appliance-level feedback. 
Using these approaches, a model is trained to predict the appliance’s total monthly consumption based on the 
monthly smart meter signal, resulting in a marked improvement over previous semi-supervised methods, and reducing the mean absolute error (MAE) of misassigned energy by an average of 70\%. 
Despite this progress, monthly-level feedback remains relatively coarse, limiting its practical utility.
Furthermore, recent studies suggest that real-time awareness of energy-intensive appliance usage can lower electricity consumption by up to 12\%~\cite{Armel2013, EDFGroupInfo, BozziCharpentier2018}, highlighting the importance of providing finer-grained details.

Non-Intrusive Load Monitoring (NILM) refers to the challenge of estimating the power consumption, pattern, or on/off state activation of individual appliances using only the aggregate smart meter household reading~\cite{DBLP:conf/eenergy/LavironDHP21, reviewnilm, themis_reviewnilm2024}.
Since the introduction of Deep-Learning (DL) for solving the NILM problem~\cite{Kelly_2015}, many different approaches have been proposed in the literature, ranging from Convolutional-based~\cite{cnn_zhang_2018, cnn1D_nilm_2019, daresnet_2019, tpnilm_2020, unet_nilm_2020, diffnilm_2023} to recent Transformer-based architectures~\cite{bert4nilm_2020, electricitynilm, energformer_2023, stnilm_2024, tsilNet_2024}.
Nevertheless, to achieve suitable performances (in terms of cost and accuracy), all of the solutions listed above shared the same setting: the DL algorithm takes as input a \emph{subsequence} of an entire household reading to estimate the individual appliances' power.
However, as shown in Figure~\ref{fig:intro}, the statistics (illustrated by the Probability Density Function) drastically vary for two subsequences extracted from the same smart meter reading.
The intermittent run of electrical appliances causes this phenomenon.
For instance, in Figure~\ref{fig:intro}, the heating system, when active, consumes a substantial amount of energy, leading to an increase in the overall power.
This change in statistical properties over time is known as the data distribution shift problem and is well known to hurt DL models' performances for time series analysis, especially in Time series forecasting (TSF)~\cite{adaptivenorm_2010, revin, Nonstationary_transformer}
In TSF, the solutions operate in a setting similar to the NILM problem, i.e., DL solutions are applied on subsequences of a long series.
Recent studies conducted in this area~\cite{revin, Nonstationary_transformer} have shown that taking into account this data distribution drift is the key to achieving accurate predictions.
To the best of our knowledge, no studies have investigated this issue in the context of energy disaggregation.

In this paper, we propose NILMFormer, a sequence-to-sequence Transformer-based architecture designed to handle the data distribution drift that occurs when operating on subsequences of an entire household consumption reading.
For this purpose, NILMFormer employs two simple but effective mechanisms, drawing inspiration from recent studies in TSF~\cite{revin, Nonstationary_transformer}.
The first one consists of making the input stationary by removing its mean and standard deviation) and passing the removed statistics information as a new token to the Transformer Block (referred to as \emph{TokenStats}).
The second one involves learning a projection of the input subsequence's statistics (mean and std) to denormalize the output signal and predict the individual appliance power (referred to as \emph{ProjStats}).
Additionally, NILMFormer employs TimeRPE, a Positional Encoding (PE) based only on the subsequence's timestamp information, which enables the model to understand time-related appliance use.
Overall, NILMFormer is significantly more accurate than current State-of-the-Art (SotA) NILM solutions, and drastically reduces the error of misassigned energy, compared to TSER methods (when used to provide per-period feedback, i.e., daily, weekly, and monthly).
NILMFormer has been successfully deployed in EDF's consumption monitoring solution \emph{Mon Suivi Conso} and currently provides millions of customers with detailed insights about their individual appliances' power consumption.

Our contributions can be summarized as follows:
    
\noindent$\bullet$ We propose NILMFormer, a sequence-to-sequence Transformer-based architecture for energy disaggregation, especially designed to handle the data distribution drift that occurs when operating on subsequences of an entire consumption series.
We also introduce TimeRPE, a PE for Transformer-based model based only on the subsequence's timestamp discrete values, which significantly boosts the performance of NILMFormer compared to standard PEs. 

\noindent$\bullet$ We evaluate NILMFormer on 4 real-world datasets against SotA NILM baselines and provide an in-depth analysis of the effectiveness of the proposed solution to handle the aforementioned data distribution drift problem.

\noindent$\bullet$ We provide insights about the deployment of our solution in EDF \emph{Mon Suivi Conso} monitoring solution, as well as highlighting the value and practicality of the solution for EDF's decision-making.

The source code of NILMFormer is available online~\cite{NilmformerCode}.

\section{Background and Related Work}
\label{sec:relatedwork}

\subsection{Non-Intrusive Load Monitoring (NILM)}
NILM~\cite{hart_nilm_1992}, also called load disaggregation, relies on identifying the individual power consumption, pattern, or on/off state activation of individual appliances using only the total aggregated load curve~\cite{themis_reviewnilm2024, eEnergy_ApplDetection, DBLP:conf/eenergy/LavironDHP21, reviewnilm}.
Early NILM solutions involved Combinatorial Optimization (CO) to estimate the proportion of total power consumption used by distinct active appliances at each time step~\cite{hart_nilm_1992}.
Later, unsupervised machine learning algorithms, such as factorial hidden Markov Models (FHMM) were used~\cite{FHMM_2011}.
NILM gained popularity in the late 2010s, following the release of smart meter datasets~\cite{Kolter2011REDDA, ukdale, refitdataset}. 
Kelly et al.~\cite{Kelly_2015} were the first to investigate DL approaches to tackle the NILM problem, and proposed 
three different architectures, including a Bidirectional Long-Short-Term Memory (BiLSTM). 
Zhang et al.~\cite{cnn_zhang_2018} proposed a Fully Convolutional Network (FCN) and introduced the sequence-to-point framework, which estimates the power consumption of individual appliances only at the middle point of an input aggregate subsequence.
However, this solution is not scalable to real-world long and large datasets currently available to suppliers, as the model needs to operate timestamp by timestamp over the entire consumption series 
to predict the individual appliances' power.
Thus, numerous sequence-to-sequence Convolutional-based architecture were later investigated, e.g., Dilated Attention ResNet~\cite{daresnet_2019}, 
Temporal Poooling~\cite{tpnilm_2020} and UNet~\cite{unet_nilm_2020}. 
With the breakthrough of the Transformer architecture~\cite{attentionisallyouneed} in Natural Language Processing, BERT4NILM~\cite{bert4nilm_2020} was proposed.
Then, variants of these architectures was proposed, such as Energformer~\cite{energformer_2023} that introduced a convolutional layer in the transformer block and replaced the original attention mechanism with its linear attention variant for efficiency. 
More recently, STNILM~\cite{stnilm_2024} is a variant of BERT4NILM that replaces the standard FFN layer with the Mixture-Of-Expert layer~\cite{switchtransformer_2022}.
In addition, hybrid architectures were recently investigated, such as BiGRU~\cite{BiGRU_2023} that mixes Convolution layers and Recurrent Units, and TSILNet~\cite{tsilNet_2024}.

All the studies mentioned above apply their solution to the subsequences of an entire household reading.
However, as pointed out in Section~\ref{sec:intro}, this leads to a large distribution drift in each subsequence, hurting model performance.
Despite the large number of studies and different proposed architectures, none of them has addressed this issue in the context of energy disaggregation. 
Moreover, and mainly due to the lack of large publicly available datasets, none of these studies provide insights into the performance of the proposed baselines in real-world scenarios involving high-consuming appliances, such as a Heating system, Water Heater, or Electric Vehicle.


\subsection{Non-Stationarity in Time Series Analysis}
DL is an established solution for tackling time series analysis tasks~\cite{tst, convtran, patchtst}, including electricity consumption series~\cite{VLDB_TransApp, themis_forecasting_2024, camal_icde, devicescope_icde}.
However, recent studies conducted in Time Series Forecasting (TSF) pointed out that the non-stationary aspect of real-world time series (e.g., the change in statistics over time) hurts model performance~\cite{adaptivenorm_2010, revin, Nonstationary_transformer}.
Indeed, similar to the NILM setting, DL methods for TSF operate on subsequences (called look-back windows); this leads to data distribution drifts over the subsequences.
To mitigate this problem, RevIN~\cite{revin}, a stationarization/destationarization scheme, is a 
solution adopted by most 
SotA TSF architectures~\cite{patchtst, samformer}.
RevIN applies a per-subsequence normalization to the input, making each series inside the network follow a similar distribution, and restores the removed statistics in the network's output.
However, this process can lead to a so-called over-stationarization, i.e., a loss of information inside the network.
The Non-Stationary Transformer framework~\cite{Nonstationary_transformer} mitigates this by combining RevIN with a handcrafted attention mechanism that feeds the removed subsequence statistical information in the network. 
Despite their success in TSF, applying such solutions directly to the NILM problem is not possible. 
Specifically, per-subsequence scaling like RevIN is not ideal for NILM since power values are closely related to specific appliances. 
Similar patterns (shapes) may be common to multiple appliances, but the maximum power differentiates them.
Moreover, restoring input statistics in the output does not align with NILM objectives, as disaggregation inherently involves a change in statistical values: an appliance's individual consumption is always lower than the aggregate input signal. 
Thus, while RevIN and similar methods are effective in TSF, they are not directly applicable to NILM.

\noindent{\bf [Data Drift Consideration in NILM]}
Mitigating the data drift when solving the NILM problem has been studied in the past.
In~\cite{nilm_drift1_rebutal}, the authors propose a framework to mitigate data drift in high-frequency measurements of electricity consumption. 
Their method uses subsequences of real, apparent, and reactive power recorded at much higher resolution than standard household smart meters, which typically measure only average real power. 
Moreover, their approach is based on classification based methods that are only able to detect whether a subsequence (window of 6sec) corresponds to an appliance’s usage (more akin to signature recognition) and cannot be used to predict and estimate appliance-specific power consumption. 
Meanwhile, Chang et al.~\cite{nilm_drift2_rebutal} present a tree-based method to adapt a trained model from one domain (eg, region or household) to another—a transfer learning perspective. 
Thus, none of these solutions tackle the intrinsic distribution drift observed in subsequences of standard smart meter data during the training phase.

\subsection{Time Series Extrinsic Regression (TSER)}
TSER solves a regression problem by learning the relationship between a time series and a continuous scalar variable, a task closely related to Time Series classification (TSC)~\cite{tser_2021}.
A recent study~\cite{tser_2021} compared the current SotA approaches for TSC applied to TSER, and showed that CNN-based neural networks (ConvNet~\cite{DBLP:journals/corr/OSheaN15}, ResNet~\cite{https://doi.org/10.48550/arxiv.1512.03385}, InceptionTime~\cite{Ismail_Fawaz_2020}), random convolutional based (Rocket)~\cite{DBLP:journals/corr/abs-1910-13051} and XGBoost~\cite{xgboost} are the current best approaches for this task.

Note that predicting the total individual appliance power consumed by an appliance over time (e.g., per day, week, or month) is a straightforward application case of TSER.

\section{Problem Definition}

A smart meter signal is a univariate time series \(x = (\boldsymbol{x}_1, ..., \boldsymbol{x}_T)\) of \(T\) timestamped power consumption readings. 
The meter reading is defined as the time difference \(\Delta_t = t_{i} - t_{i-1}\) between two consecutive timestamps \(t_i\). 
Each element \(\boldsymbol{x}_t\), typically measured in Watts or Watt-hours, represents either the actual power at time \(t\) or the average power over the interval \(\Delta_t\).



\noindent{\textbf{[Energy Disaggregation]}} The aggregate power consumption is defined as the sum of the $N$ individual appliance power signal $a_1(t), a_2(t), \ldots, a_N(t)$ that run simultaneously plus some noise $\epsilon(t)$, accounting for measurement errors. 
Formally, it is defined as:
{\small
\begin{equation}
    x(t)=\sum^{N}_{i=0} a_i(t) + \epsilon(t)
\end{equation}
}
where $x(t)$ is the total power consumption measured by the main meter at timestep $t$; $N$ is the total number of appliances connected to the smart meter; and $\epsilon(t)$ is defined as the noise or the measurement error at timestep $t$.
The NILM challenge relies on accurately decomposing $x(t)$ to retrieve the $a_i(t)$ components.

\section{Proposed Approach}
\label{sec:proposedapproach}

\begin{figure}
    \centering
    \includegraphics[width=1\linewidth]{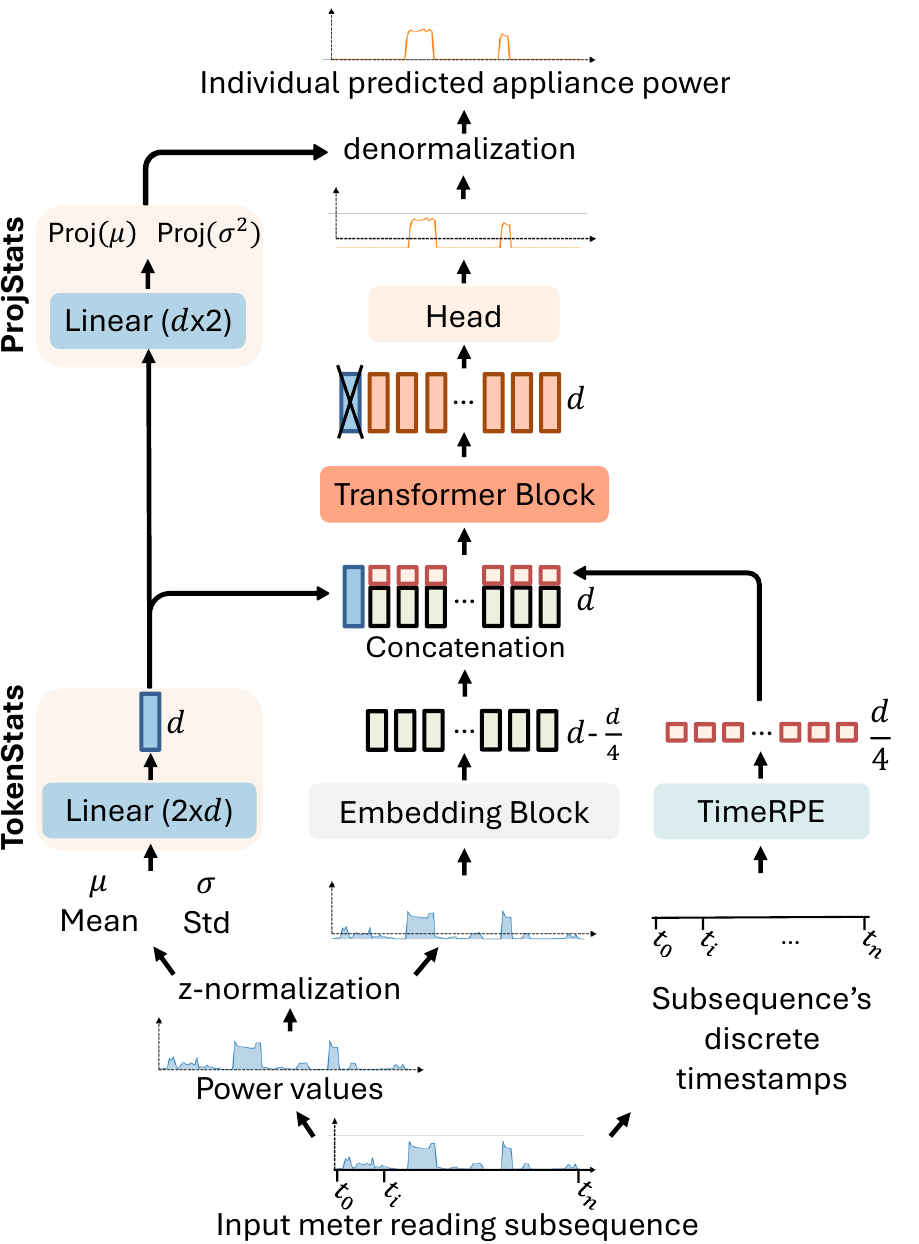}
    \caption{Overview of the NILMFormer architecture.}
    \label{fig:nilmformer} 
\end{figure}

The Transformer architecture has demonstrated good performance when applied to energy disaggregation~\cite{bert4nilm_2020, energformer_2023, stnilm_2024}, but current approaches do not consider the non-stationary aspect of real-world smart meter data.
We propose NILMFormer as a sequence-to-sequen\-ce Transformer-based architecture designed to handle this phenomenon.
NILMFormer operates by first stationarizing the input subsequence by subtracting its mean and standard deviation.
While the normalized subsequence is passed through a robust convolutional block that serves as a feature extractor, the removed statistics are linearly projected in a vector (referred to as \emph{TokenStats}), and the proposed TimeRPE module uses the timestamps to compute a positional encoding matrix.
These features are concatenated and fed into the Transformer block, followed by a simple Head to obtain a 1D sequence of values.
The final step consists of linearly projecting back the \emph{TokenStats} (referred to as \emph{ProjStats}) to 2 scalar values that are then used to denormalize the output, providing the final individual appliance consumption.

Overall, NILMFormer first splits and encodes separately the shape, the temporal information, and the intrinsic statistics of the subsequences, which are then mixed back together in the Transformer block.
In addition, the output prediction is refined through the linear transformation of the input series statistics, accounting for the loss of power when disaggregating the signal.

\subsection{NILMFormer Architecture}

As depicted in Figure~\ref{fig:nilmformer}, NILMFormer results in an encoder that takes as input a subsequence of an entire household smart meter reading $x = (\boldsymbol{x}_1,  ..., \boldsymbol{x}_n)$ and outputs the individual consumption $a = (\boldsymbol{a}_1,  ..., \boldsymbol{a}_n)$ for an appliance.
The workflow unfolds in the following steps.

\noindent \textbf{Step 1: Input subsequence stationarization.} The input power subsequence $x^{1 \times n}$ is first z-normalized.
More specifically, we first extract the mean and standard deviation of the sequence as $\mu = \frac{1}{n} \sum_{i=0}^n x_i$ and $\sigma = \sqrt{\frac{1}{n} \sum_{i=0}^n (x_i - \mu)^2}$, respectively.
Then, we remove the extracted mean to each value $\boldsymbol{x}_i$ and divide them by the standard deviation as $\tilde{x} = \frac{x-\mu}{\sigma + \epsilon}$, such that the mean of the subsequences become 0 and the standard deviation 1.

\noindent \textbf{Step 2: Tokenization.} The mean $\mu$ and standard deviation $\sigma$ values are projected using a learnable linear layer in a $d$-dimensional vector, referred to as \emph{TokenStats}.
In addition, the z-normalized subsequence $\tilde{x}^{1 \times n}$ is passed through the embedding block, used to extract local features.
This block is composed of several convolutional layers using $d-\frac{d}{4}$ filters (see Section~\ref{sec:embddingblock} for details), which output a features map $z^{d-\frac{d}{4} \times n}$.
In parallel, the positional encoding matrix $PE^{\frac{d}{4} \times n}$ is computed according to the subsequence's discrete timestamp information using the TimeRPE module (detailed in Section~\ref{sec:trpe}).

\noindent \textbf{Step 3: Features mix.} The \emph{TokenStats} and the positional encoding matrix are concatenated to the extracted subsequence's features maps, resulting in a new features map $\hat{z}^{(n+1) \times d}$.
More specifically, the positional encoding matrix is concatenated along the inner dimension $d$, and the \emph{TokenStats} is concatenated along the time dimension $n$ (it can be viewed as adding a new token).
Then, the obtained matrix $\hat{z}$ is passed through the Transformer Block (see Section~\ref{sec:transformerblock} for details) that is used to mix the different information and learn long-range dependencies.
Then, the first token of the output representation $\hat{z}^{(n+1) \times d}$, corresponding to the \emph{TokenStats}, is removed to obtain a feature map matching the input subsequence length, $z^{n \times d}$.
Finally, $z^{n \times d}$ is processed through the output Head, consisting of a 1D convolutional layer, which maps the latent representation back to a 1D series representation, $\tilde{a}^{1 \times n}$.

\noindent \textbf{Step 4: Output de-stationarization.} The \emph{TokenStats} is projected back using a learnable linear layer ($d \times 2$) that provides two scalar values $Proj(\mu)$ and $Proj(\sigma)$, referred to as \emph{ProjStats}.
These values are used as a new mean and standard deviation to denormalize the output $\tilde{a}^{1 \times n}$ and obtain the final prediction as: $a = \tilde{a} * Proj(\sigma) + Proj(\mu)$.
The subsequence $a$ is the individual appliance power.


\subsubsection{Embedding Block} 
\label{sec:embddingblock}

As depicted in Figure~\ref{fig:nilmformerparts}(b), the Embedding Block results in 4 stacked convolutional Residual Units (ResUnit), each one composed of a convolutional layer, a GeLU activation function~\cite{gelu}, and a BatchNormalization layer~\cite{https://doi.org/10.48550/arxiv.1502.03167}.
Not that a residual connection is used between each ResUnit, and a stride parameter of 1 is employed in each convolutional filter to keep the time dimension unchanged.
For each Residual Unit $i=1,...,4$, a dilation parameter $d=2^i$ that exponentially increases according to the ResUnit's depth is employed. 
We motivate the choice of using this feature extractor in Section~\ref{sec:ablation}.

\subsubsection{Transformer Block} 
\label{sec:transformerblock}

The core of NILMFormer relies on a block composed of $N$ stacked Transformer layers.
Each Transformer layer is made of the following elements (cf. Figure~\ref{fig:nilmformerparts}(a)): a normalization layer, a Multi-Head Diagonally Masked Self-Attention mechanism (Multi-Head DMSA), a second normalization layer, and a Positional Feed-Forward Network~\cite{attentionisallyouneed} (PFFN). 
We use a Multi-Head DMSA instead of the original Attention Mechanism, as it is more effective when applied to electricity consumption series analysis~\cite{VLDB_TransApp}.
In addition, we introduce residual connections after the Multi-Head DMSA and the PFFN, and the use of a Dropout parameter.

\subsubsection{Timestamp Related Positional Encoding (TimeRPE)} 
\label{sec:trpe}

\begin{figure}
    \centering
    \includegraphics[width=1\linewidth]{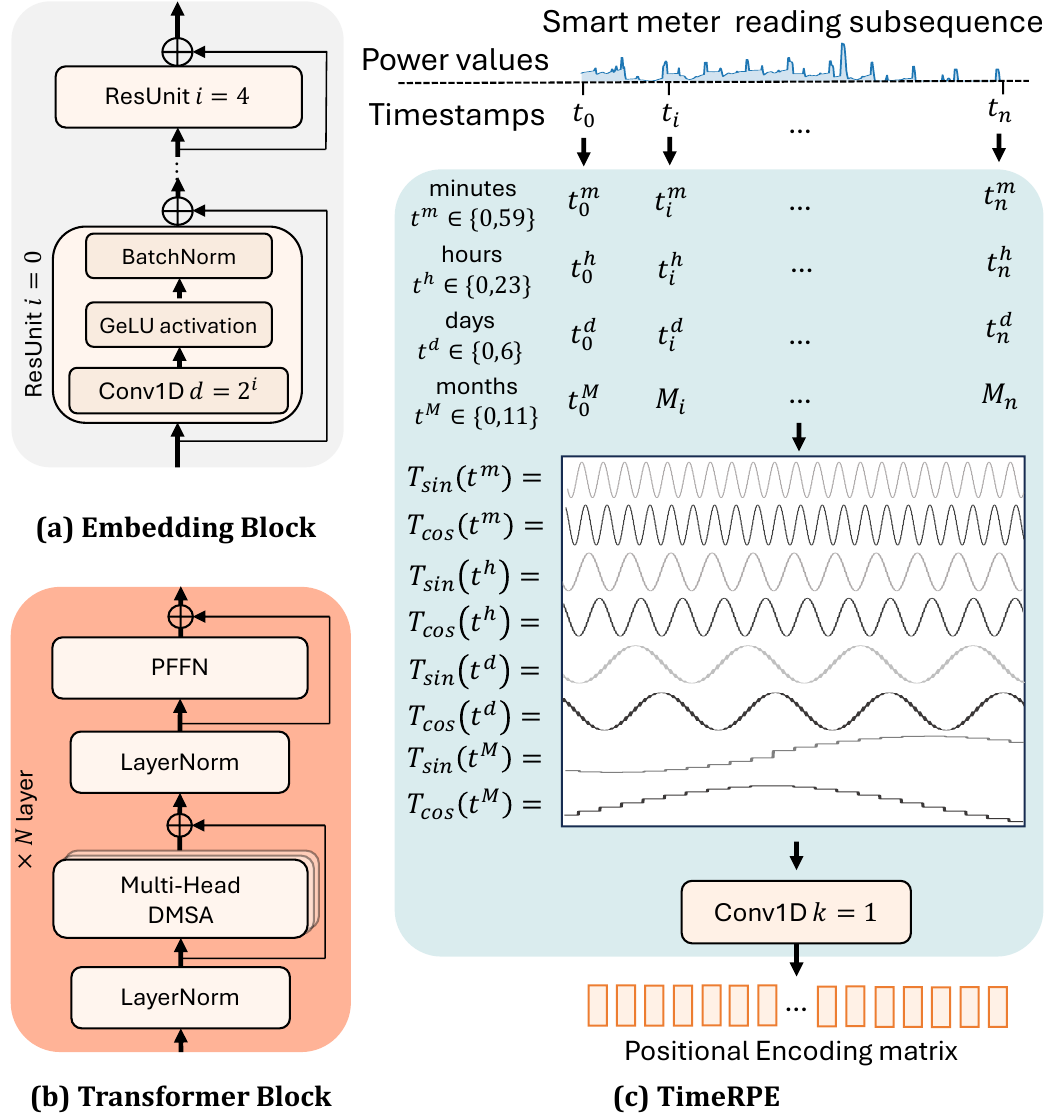}
    \caption{NILMFormer's architecture parts detail: (a) Embedding Block; (b) Transformer Block; (c) TimeRPE module.}
    \label{fig:nilmformerparts} 
\end{figure}

The Transformer architecture does not inherently understand sequence order due to its self-attention mechanisms, which are permutation invariant. 
Therefore, Positional Encoding (PE) is mandatory to provide this context, allowing the model to consider the position of each token in a sequence~\cite{attentionisallyouneed}. 
Fixed sinusoidal or fully learnable PEs are commonly used in most current Transformer-based architectures for time series analysis~\cite{patchtst}, including those proposed for energy disaggregation~\cite{bert4nilm_2020, transnilm_2022}. 
This kind of PE consists of adding a matrix of fixed or learnable weight on the extracted features before the Transformer block.
However, these PEs only help the model understand local context information (i.e., the given order of the tokens in the sequence) and do not provide any information about the global context when operating on subsequences of a longer series. 
In the context of NILM, appliance use is often related to specific periods (e.g., dishwashers running after mealtimes, electric vehicles charging at night, or on weekends). 
Moreover, detailed timestamp information is always available in real-world NILM applications.
Thus, using a PE based on timestamp information can help the model better understand the recurrent use of appliances. 
Timestamp-based PEs have been briefly investigated for time series forecasting~\cite{informer} but were always combined with a fixed or learnable PE and directly added to the extracted features.

Therefore, we proposed the Timestamps Related Positional Encoding (TimeRPE), a Positional Encoding based only on the discrete timestamp values extracted from the input subsequences.
The TimeRPE module, depicted in Figure~\ref{fig:nilmformerparts} (c), takes as input the timestamps information $t$ from the input subsequences, decomposes it such as minutes $t^m$, hours $t^h$, days $t^d$, and months $t^M$, and project them in a sinusoidal basis, as:
    $T_{\sin}(t_i) = \sin \left(\frac{2 \pi t^j_i}{p^j} \right) \quad \text{and} \quad
    T_{\cos}(t_i) = \cos \left(\frac{2 \pi t^j_i}{p^j} \right)$,
with $j \in \{m, h, d, M\}$ and $\{p^m=59, p^h=23, p^d=6, p^M=11\}$ corresponding to the set of max possible discrete timestamp variable.
Afterward, the obtained representation is projected in a higher-dimensional space using a 1D convolution layer with a kernel of size 1.
We evaluate the effectiveness of TimeRPE against various standard PE used in time series Transformers in Section~\ref{sec:ablation}.

\section{Experimental Evaluation}
\label{sec:expsetup} 

All experiments are performed on a 
cluster with 2 Intel Xeon Platinum 8260 CPUs, 384GB RAM, and 4 NVidia Tesla V100 GPUs with 32GB RAM.
The source code~\cite{NilmformerCode} is in Python v3.10, and the core of NILMFormer in 
PyTorch v2.5.1~\cite{pytorch}.

\subsection{Datasets}
\label{sec:datasets} 

\begin{table}[tb]
\caption{\label{table:detaildataset} Dataset characteristics and 
parameters}
\begin{adjustbox}{width=\columnwidth,center}
\begin{tabular}{c|c|c|c|c}
    \toprule
    {\bf Datasets} & UKDALE & REFIT & EDF 1 &  EDF 2  \\
    \midrule
    {\bf Nb. Houses} & 5 & 20 & 369 & 24  \\
    \midrule
    {\bf Avg. recording time} & 223days & 150days & 3.7years & 1year \\
    \midrule
    {\bf Sampling rate} & 1min & 1min & 10min & 30min  \\
    \midrule
    {\bf Max. power limit ($\tau_ {max}$)} & 6000 & 10000 & 24000 & 13000  \\
    \midrule
    {\bf Max. ffill} & 3min & 3min & 30min & 1h30  \\
    \midrule
    \multirow{5}{*}{{\bf Appliances}} 
    & Dishwasher & Dishwasher & Central Heater & Electric Vehicle  \\
    & Washing Machine & Washing Machine & Heatpump &  \\
    & Microwave & Microwave & Water Heater &  \\
    & Kettle & Kettle & WhiteUsages & \\
    & Fridge & & &  \\
    \bottomrule
\end{tabular}
\end{adjustbox}
\end{table}

We use 4 different datasets (see Table~\ref{table:detaildataset}) that provide the total power consumed by the house, recorded by a smart meter, and the individual load power measurement for a set of appliances.

\subsubsection{Public Datasets} UKDALE~\cite{ukdale} and REFIT~\cite{refitdataset} are two well-known 
datasets used in many research papers to assess the performance of NILM approaches~\cite{Kelly_2015, cnn_zhang_2018, bert4nilm_2020, diffnilm_2023, tsilNet_2024}.
The two datasets contain high-frequency sampled data collected from small groups of houses in the UK and focus on small appliances. 

\import{./}{resultstablesmall.tex}

\subsubsection{EDF Datasets} 
 
We also use 2 private (EDF) datasets that include modern appliances not included in the public datasets, such as Heater systems, Water Heaters and Electric Vehicle chargers.

\noindent{\textbf{[EDF Dataset 1]}} It contains data from 358 houses in France between 
2010 and 
2014.
Houses were recorded for an average of 1357 days (shortest: 501 days; longest: 1504 days).
Houses were monitored with smart meters that recorded the aggregate main power, as well as the consumption of individual appliances at 10min intervals.
Diverse systems, including the central heating system, the heat pump, and the water heater, have been monitored directly through the electrical panel.
In addition, other appliances, such as the dishwasher, washing machine, and dryer, were monitored using meter sockets.
The signals collected from these appliances have been grouped in one channel called "White Appliances". 

\noindent{\textbf{[EDF Dataset 2]}} It contains data from 24 houses in France from July 2022 to February 2024.
Data were recorded for an average of 397 days (shortest: 175 days; longest: 587 days).
Houses were monitored with individual smart meters that recorded the main aggregate power of the house and a clamp meter that recorded the power consumed by an electric vehicle's charger.
Aggregate main power and electric vehicle recharge were sampled 30min intervals.

\subsubsection{Data processing}

According to the parameters reported in Table~\ref{table:detaildataset}, we resampled and readjusted recorded values to round timestamps by averaging the power consumed during the interval $\Delta_t$, we forward-filled the missing values, and we clipped the consumption values between 0 and a maximum power threshold.

In this study, we evaluate the model's performance based on real-world scenarios using unseen data from different houses within the same dataset~\cite{themis_reviewnilm2024}. 
Distinct houses were used for training and evaluation to ensure robust performance assessment. For the UKDALE dataset, we utilized houses 1, 3, 4, and 5 for training and house 2 for testing.
This selection was made because only houses 1 and 2 contained all the appliances. 
Note that for UKDALE, 80\% of the data from house 1 was used for training, while 20\% was used as a validation set to prevent overfitting.

For the REFIT and EDF1 datasets, which contain more houses, 2 houses were reserved for testing, 1 house for validation, and the remaining houses were used for training. 
For 
EDF2, which contains more than 350 houses, 70\% of the houses were used for training, 10\% for validation, and the remaining 20\% for evaluation.

For training and evaluating the model, the entire recorded consumption data of the different houses was sliced into subsequences using a non-overlapping window of length $w$. 
Subsequences containing any remaining missing values were discarded. 
To ensure training stability, we scaled the data to a range between 0 and 1 by dividing the consumption values (both aggregate and individual appliance power) by the maximum power threshold $\tau_{max}$ for each dataset as reported in Table~\ref{table:detaildataset}. 
Consequently, before evaluating the models, we denormalized the data by multiplying it by $\tau_{max}$.

\subsection{Evaluation Pipeline}
\label{sec:pipeline}

\begin{figure*}[tb]
    \centering
    \includegraphics[width=1\linewidth]{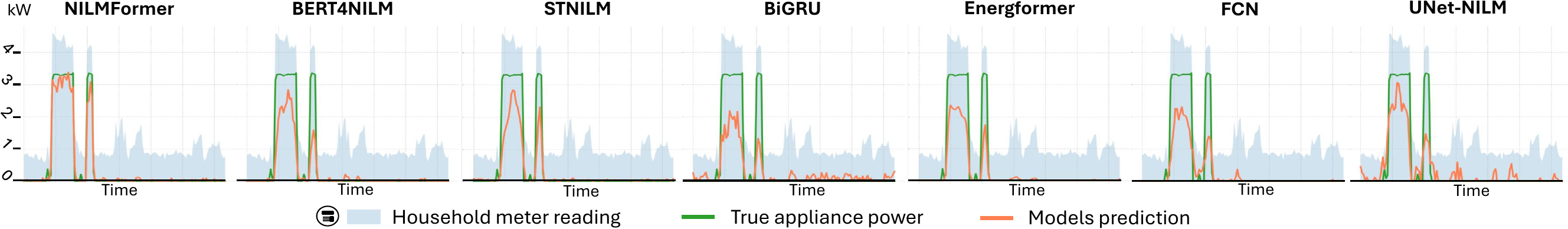}
    \caption{Qualitative disaggregation results for Electric Vehicle (EDF1) on a sample example for the 7 best baselines. ($w=128$).}
    \label{fig:SampleResults} 
    \vspace{0.4cm}
\end{figure*}

We compare our solution against several SotA sequence-to-sequence NILM solutions.
We include two recurrent-based architectures, BiLSTM~\cite{Kelly_2015} and BiGRU~\cite{BiGRU_2023} that combine Convolution and Recurrent layers; three convolutional-based baselines, FCN~\cite{cnn_zhang_2018}, UNet-NILM~\cite{unet_2015} and DAResNet~\cite{daresnet_2019}; as well as a recent diffusion-based proposed approach, DiffNILM~\cite{diffnilm_2023}.
In addition, we include 4 Tranfor\-mer-based baselines, BERT4NILM~\cite{bert4nilm_2020}, Energformer~\cite{energformer_2023}, STNILM~\cite{stnilm_2024}, nd TSILNet~\cite{tsilNet_2024}, which integrates both Transformer and recurrent layers.
We used the default parameters provided by the authors and trained all the models using the Mean Squared Error loss.


\subsubsection{Window length sensitivity}
We evaluate each baseline using different subsequences windows length $w$ to assess the sensitivity of the results on this parameter.
We experimented with windows length $w = \{128, 256, 512\}$ to standardize results across datasets. 
We also considered using window lengths corresponding to specific periods based on the sampling rate but observed no significant changes in the results.

\subsubsection{Evaluation metrics}

We assess the energy disaggregation quality performance of the different baselines using 2 metrics.
For each metric, $T$ represents the total number of intervals while $y_t$ is the true and $\hat{y}_t$ is the predicted power usage of an appliance.
The first metric is the standard Mean Absolute Error (MAE):
$MAE = \frac{1}{T} \sum_{t=1}^{T} |\hat{y}_t - y_t|$.
The second metric is the Matching Ratio (MR), based on the overlapping rate of true and estimated prediction, and stated to be the best overall indicator performance~\cite{Mayhorn2016LoadDT}:
$MR = \frac{\sum_{t=1}^{N} min(\hat{y}_t, y_t)}{\sum_{t=1}^{N} max(\hat{y}_t, y_t)}$.

\subsection{Results}
\label{sec:results}


Table~\ref{table:resultssmall} lists the 
results for the 4 datasets and each different appliance energy disaggregation case and subsequence window length.
First, we note that NILMFormer outperforms other solutions overall for the 2 metrics (avg. rank per metric and avg. total Rank).
More specifically, we note an improvement of over 15\% in terms of MAE and 22\% in MR (on average across all the datasets and cases) compared to the second-best model for each metric, STNILM and BERT4NILM, respectively.
NILMFormer is outperformed only on the Fridge disaggregation case on the UKDALE dataset by UNet-NILM.
This is because the fridge is always ON and present in all houses. 
As a result, the scheme adopted to address the non-stationary nature of the data caused the model to struggle when isolating the constant, baseline consumption that the fridge represents.
We also provide in Figure~\ref{fig:SampleResults} an example of the disaggregation quality of the 7 best baselines on the Electric Vehicle case (EDF1).

\subsection{Impact of Design Choices on Performance}
\label{sec:ablation}

We perform a detailed evaluation to assess the performance of different key parts of NILMFormer.
We first study the effectiveness of the two mechanisms proposed to mitigate the non-stationary aspect of the data (i.e., the \emph{TokenStats} and the \emph{ProjStats}).
Then, we assess the proposed TimeRPE's performance against other PEs usually used in Transformers for time series analysis.

We utilize a Critical Difference Diagram (CD-Diagram) to compare the performance, computed based on the rank of the different variants. 
This method, proposed in~\cite{CDDiagramRank}, involves performing a Friedman test followed by a post-hoc Wilcoxon test on the calculated ranks, according to significance level $\alpha = 0.1$.
The rank is given by computing an overall score according to the two disaggregation metrics (MAE and MR) averaged across all the datasets, cases, and window lengths, previously normalized to the same range (between 0 and 1).
In addition, we note that bold lines in CD-diagrams indicate insignificant differences between the connected methods.

\subsubsection{Non-stationarity consideration} We assess the influence of the two mechanisms adopted to mitigate the non-stationary aspect of the data.
More precisely, we compared the following options: 
(\textbf{None}): not considering any mechanism to handle the non-stationary aspect of the data (as the previously proposed NILM approach);
(\textbf{RevIN}): applying the RevIN~\cite{revin} per subsequences normalization/denormalization scheme to the input/output without propagating any information inside the network, i.e., the approach currently adopted in most of TSF architecture~\cite{patchtst, samformer};
(\textbf{w/ TokenStats}): propagating only the \emph{TokenStats} inside the Transformer block, i.e., the extracted input statistics $\mu$ and $\sigma$ are re-used to denormalize the output,
(\textbf{w/ ProjStats}): using only the learnable projection of the extracted input statistics $\mu$ and $\sigma$ to denormalize the output without propagating information in the Transformer Block; 
and,
(\textbf{w/ TokenStats and w/ ProjStats}); the final proposed mechanism include in NILMFormer.
The results (cf. Figure~\ref{fig:AbCDAll}(a)) demonstrate that applying only the RevIN scheme leads to worse performance than using the proposed architecture without any mitigation of the non-stationarity effect and, thus, confirms that level information is crucial for NILM.
In addition, omitting either \emph{TokenStats} or \emph{ProjStats} results in a performance drop, confirming the essential role of both.

\subsubsection{Positional Encoding}
\begin{figure}[tb]
    \centering
    \includegraphics[width=1\linewidth]{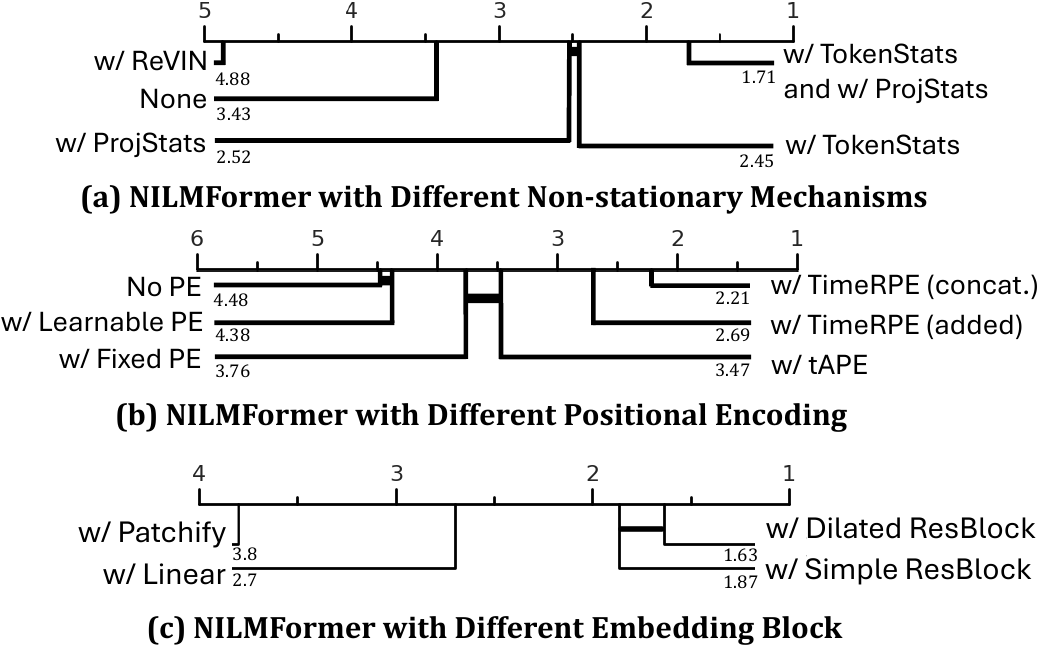}
    \caption{CD-diagram of the average rank (avg. of the MAE and MR across all datasets, cases, and window lengths) to study the impact of: (a) the proposed mechanisms to mitigate the non-stationary aspect of the data, (b) different PE, and (c) different embedding block.}
    \label{fig:AbCDAll} 
    \vspace*{0.4cm}
\end{figure}

We evaluate the effectiveness of the proposed TimeRPE against usual standard PE methods used in the NILM literature.
More specifically, we investigate the following options: (1) removing the PE (\textbf{NoPE}); (2) concatenating or adding the TimeRPE's PE matrix to the embedding features; (3) replacing TimeRPE by (\textbf{Fixed}) the original PE proposed for Transformer in~\cite{attentionisallyouneed}); (\textbf{tAPE}) a fixed PE proposed recently, especially for time series Transformer~\cite{convtran}; (\textbf{Learnable}) a fully learnable one, as proposed for time series analysis in~\cite{tst}, and used in the NILM literature~\cite{bert4nilm_2020, electricitynilm, midformer_2022}.
The critical diagram in Figure~\ref{fig:AbCDAll}(b) shows the significant superiority of TimeRPE over other PE options.
Moreover, we noticed that concatenating the PE information with the features extracted from the Embedding Block instead of adding it lead to significantly better performance.
We assume that concatenating the PE instead of adding lead helps the model to differentiate the two pieces of information.

\begin{figure}[tb]
    \centering
    \includegraphics[width=0.72\linewidth]{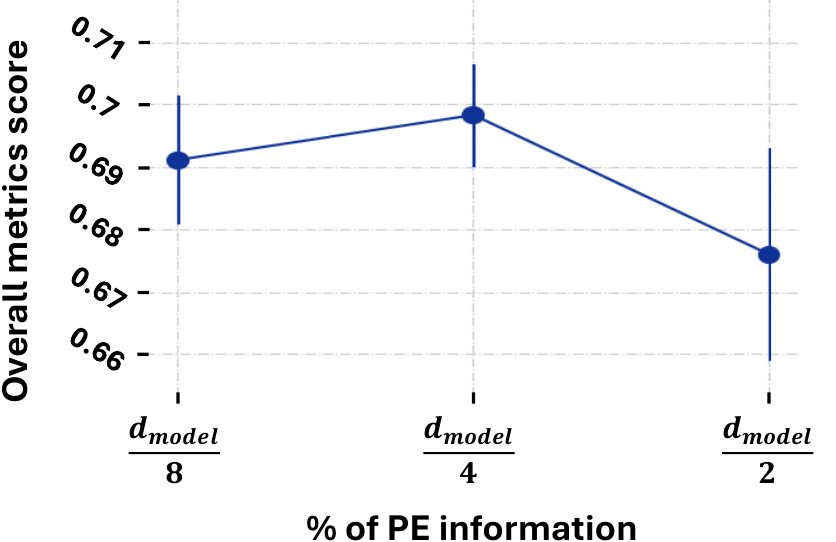}
    \caption{Overall metrics score (avg. of the MAE and MR for $w=256$, over all the datasets and cases) by varying the ratio of positional encoded information (given by TimeRPE) according to the inner model dimension ($d_{model}=128$).}
    \label{fig:pepercentageimpact} 
    \vspace{0.4cm}
\end{figure}

\subsubsection{Embedding Block}
We evaluate the impact of different embedding blocks for extracting features from the aggregate power signal.
More specifically, we investigate replacing the proposed Dilated Residual Embedding Block by: (\textbf{Linear}) a simple linear embedding layer that maps each time step of the input sequence model; (\textbf{Patchify}) a patching embedding, an approach used by numerous Transformer for time series analysis~\cite{midformer_2022, patchtst}, that involves dividing the input series in patches (i.e., subsequences); using a convolutional layer with stride and kernel of the patch length; and
(\textbf{ResBlock}) a simple Residual Convolution Block without dilation (with k=3).
The results (cf. Figure~\ref{fig:AbCDAll}(c)) demonstrates the necessity of the convolution layers to extract localized feature patterns.
Using the proposed Embedding Block leads to a slight (but not significant) increase compared to the simple ConvBlock.
Note that using the patch embedding leads to the worst results, suggesting that this type of embedding does not suit sequence-to-sequence NILM solutions.

\subsubsection{Impact of Positional Encoding Ratio on Model Performance}

As detailed in Section~\ref{sec:proposedapproach}, NILMFormer concatenates the positional-encoding matrix produced by TimeRPE with the feature vectors generated by the Embedding Block along the inner model dimension $d$.  
To quantify how much representational budget should be allocated to positional cues, we train NILMFormer while varying the share of channels reserved for TimeRPE (using window size $w=256$ and $d=128$).
Specifically, we examine three ratios, $\bigl\{\tfrac{d}{8},\,\tfrac{d}{4},\,\tfrac{d}{2}\bigr\}$, corresponding to dedicating $12.5\%$, $25\%$, and $50\%$ of the channels to positional information.  

The results in Figure~\ref{fig:pepercentageimpact} reveal that allocating one quarter of the channels to positional encoding ($\tfrac{d}{4}$) offers the best trade-off, delivering the highest disaggregation accuracy.  
Increasing the share to $\tfrac{d}{2}$ brings no further gains, while reducing it to $\tfrac{d}{8}$ only slightly affects performance.

\section{Deployed Solution}
\label{sec:edfscenario}

\begin{figure*}[tb]
    \centering
    \includegraphics[width=1\linewidth]{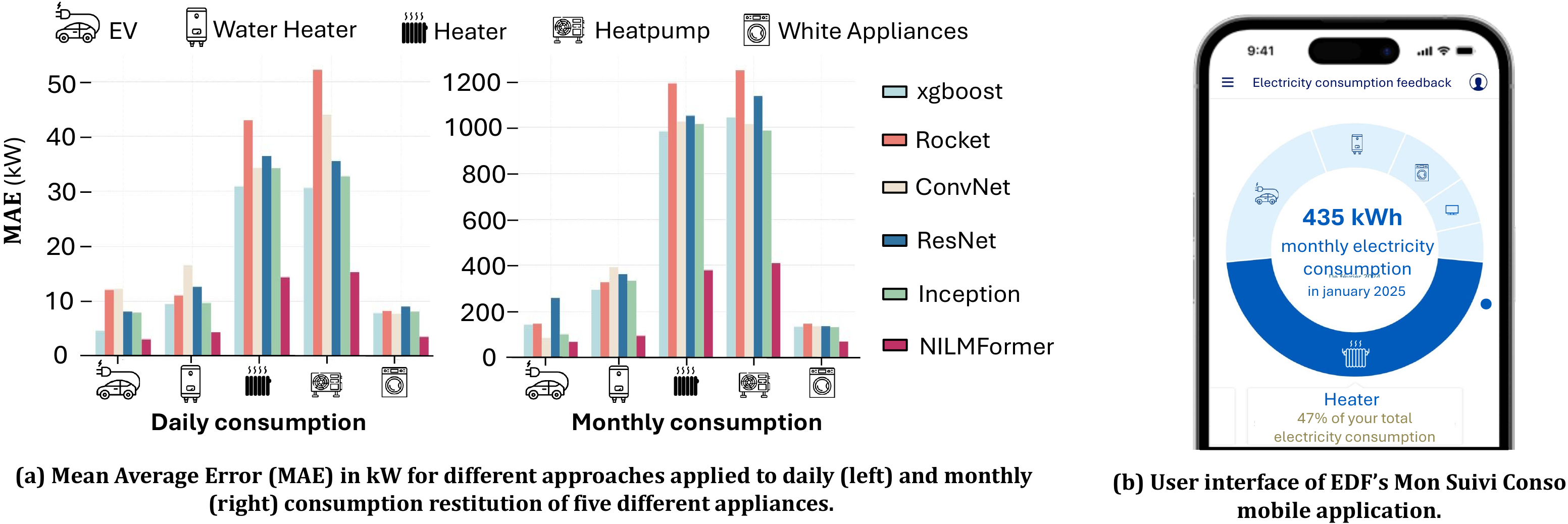}
    \caption{(a) Results comparison for per-period energy estimation; (b) Example of feedback available to a client through the user interface of EDF's \emph{Mon Suivi Conso} mobile application.}
    \label{fig:ResultsApplicationAndUI} 
    \vspace*{0.3cm}
\end{figure*}

Since 2015, EDF has offered to its clients a consumption monitoring service called \emph{Mon Suivi Conso}~\cite{SuiviConsoEDF}, enabling clients to track their consumption.
The service is accessible via the web and a mobile app.
A new feature was released in 2018 to provide the clients with an estimation of their annual individual appliance consumption.
The backbone algorithm relied on semi-supervised statistical methods that used the customers’ static information~\cite{patentEDF1, patentEDF2}. 
However, recent user feedback indicated a growing demand for more granular and personalized insights, consumption estimates down to daily frequency, and percentage-based cost breakdowns per appliance~\cite{EDFprivatecomunicationdatanumia}. 
In response, EDF recently explored the use of TSER~\cite{tser_2021} to infer monthly appliance-level consumption. 
These approaches yielded an improvement over the original semi-supervised methods employed in \emph{Mon Suivi Conso}, reducing the mean absolute error (MAE) of monthly misassigned consumption 
across various appliances (e.g., heaters, water heaters, and electric vehicles).
Despite these advances, monthly-level feedback remains relatively coarse, limiting its practical value for end-users. 
Consequently, NILM-based algorithms, which can disaggregate consumption at a per-timestamp level, offer a promising alternative for replacing and enhancing the current individual appliance consumption feedback feature.

\noindent\textbf{[NILMFormer Applied to Daily and Monthly Feedback]}
Employing NILMFormer for per-period feedback is straightforward, as the smart meters in France collect data at half-hour intervals.
Thus, to provide the individual appliance consumption over a period of time, we adapted NILMFormer as follows:

\noindent{1.} The electricity consumption series of a client is sliced into subsequences using a tumbling (non-overlapping) window of size $w$.

\noindent{2.} Each subsequence is then passed to a NILMFormer instance trained to disaggregate the signal for a specific appliance $a$.

\noindent{3.} The predictions are then concatenated.

\noindent{4.} Finally, the appliance predicted power consumption is returned, summing over the period of interest (day, week, or month).

\noindent{\bf [Comparison to TSER Approaches]} We experimentally evaluate the performance of our approach against the 5 best SotA TSER baselines reported in~\cite{tser_2021}, including XGBoost~\cite{xgboost}, Rocket~\cite{DBLP:journals/corr/abs-1910-13051} (a random convolutional-based regressor), and 3 convolutional-based DL regressors (ConvNet, ResNet, and InceptionTime).
We consider two settings: predicting the per-day and per-month appliance consumption.
To train the TSER baselines, we preprocessed the datasets (EDF1 and EDF2) by slicing the entire consumption into subsequences (day or month), and by summing the total consumption of the appliance over that period to obtain the label.
Note that all baselines were initialized using default parameters, and DL-based methods were trained using the Mean Squared Error Loss.

To evaluate our approach, we reused the different instances of NILMFormer trained on timestamp energy disaggregation in Section~\ref{sec:results} and applied the framework described above.

\paragraph{Results} The results (cf. Figure~\ref{fig:ResultsApplicationAndUI}(a)) demonstrate the superiority of NILMFormer, which significantly outperforms TSER baselines applied to daily and monthly appliance consumption prediction. 
On average, across the five appliances evaluated, NILMFormer achieves a 51\% reduction in MAE for daily consumption compared to the second-best baseline (XGBoost), and a 151\% reduction in MAE for monthly consumption compared to the second-best baseline (Inception). 
These findings highlight the effectiveness of our approach for deployment in \emph{Mon Suivi Conso}, providing substantially more accurate feedback to clients than previous approaches.

\subsection{Deployment Insights and Performance}

DATANUMIA, an EDF subsidiary, oversees the deployment and integration of NILMFormer into \emph{Mon Suivi Conso}, gradually replacing the existing solution for individual appliance consumption feedback. 
The final delivered algorithm uses a separate model for each appliance, enabling the solution to predict consumption for each appliance independently. 
These individual predictions are then aggregated to produce a final output showing the percentage of each appliance’s consumption relative to the total (see Figure~\ref{fig:ResultsApplicationAndUI}(b)). 
Users can also track their electricity usage according to different time intervals: daily, monthly, and annual periods, both in kilowatt-hours and euros (based on their contracted rates). 

\noindent \textbf{[Infrastructure and Performance]} The deployed solution is hosted on Amazon Elastic Compute Cloud (EC2) using 6 \emph{m6i.8xlarge} instances, each providing 32 vCPUs. 
The individual appliance consumption estimation runs weekly on the entire customer base that has consented to data analysis—around 3.6 million customers. 
Post-launch performance evaluations shows that the system currently handles 100 customer requests per second, processing the entire dataset in approximately 11 hours.

\noindent \textbf{[User Engagement]} During the last quarter of 2024, operational statistics indicate 7 million visits to the consumption monitoring solution that uses our approach, with 2.2 million visits to the individual appliance consumption feedback through the web interface. 
For the mobile application, out of 7 million visits to this application, 6.2 million visits focused on the per-appliance consumption feature, demonstrating considerable interest in the information that our solution provides.

\noindent \textbf{[Decision Making]} EDF recently applied this solution to a subset of consenting consumers on La Réunion~\cite{edf_sarz_la_kaz}, focusing on electric vehicle (EV) charging habits, and the impact of different off-peak charging systems (e.g., smart plugs versus dedicated charging stations) on overall consumption. 
Through NILMFormer, it was possible to pinpoint exact EV charging times and calculate the associated total consumption. 
The results showed notable benefits for customers using controlled charging stations, offering improved insights into peak/off-peak ratios, and demonstrating the value of NILMFormer in enhancing data-driven decision-making for EDF.

\section{Conclusions}

We proposed NILMFormer, a DL sequence-to-sequence Transformer for energy disaggregation, designed to address the data distribution drift problem that occurs when operating on subsequences of consumption series.
NILMFormer employs a stationarization/de-stationarization scheme tailored to the NILM problem and uses TimeRPE, a novel PE based only on the subsequence's timestamp information.
The results show that NILMFormer significantly outperforms current SotA NILM solutions on 4 different datasets. 
NILMFormer outperforms the previous method for individual appliance per period feedback.
Our solution has been successfully deployed as the backbone algorithm for EDF's consumption monitoring service,  
delivering detailed insights to millions of customers. 

\begin{acks}
Supported by EDF R\&D, ANRT French program, and EU Horizon projects AI4Europe ($101070000$), TwinODIS ($101160009$), ARMADA ($101168951$), DataGEMS ($101188416$), RECITALS ($101168490$), and by $Y \Pi AI \Theta A$ \& NextGenerationEU project HARSH ($Y\Pi 3TA-0560901$).
\end{acks}


\bibliographystyle{ACM-Reference-Format}
\balance
\bibliography{bibliography}


\newpage
\appendix
\section{NILMFormer Architecture Details}

This section provides additional details and insights about the NILMFormer architecture.

\subsection{Hyperparameters}

We report in Table~\ref{table:hyperparameters} the list of hyperparameters used in NILMFormer for our experiments.

\subsection{Hyperparameter Impact}

We experimentally evaluate the impact of two main NILMFormer hyperparameters (the inner dimension $d$ and the number of Transformer layers) on the disaggregation performances and the complexity (in terms of the number of trainable parameters).
More specifically, we trained and evaluated NILMFormer in the setting described in Section~\ref{sec:expsetup} (for all the datasets, cases, and with $w=256$) for the following inner dimension: $d = \{64, 96, 128, 256\}$, and the following number of Transformer layers $n_l = \{1, 3, 5, 7\}$.

\begin{figure}[tb]
    \centering
    \includegraphics[width=1\linewidth]{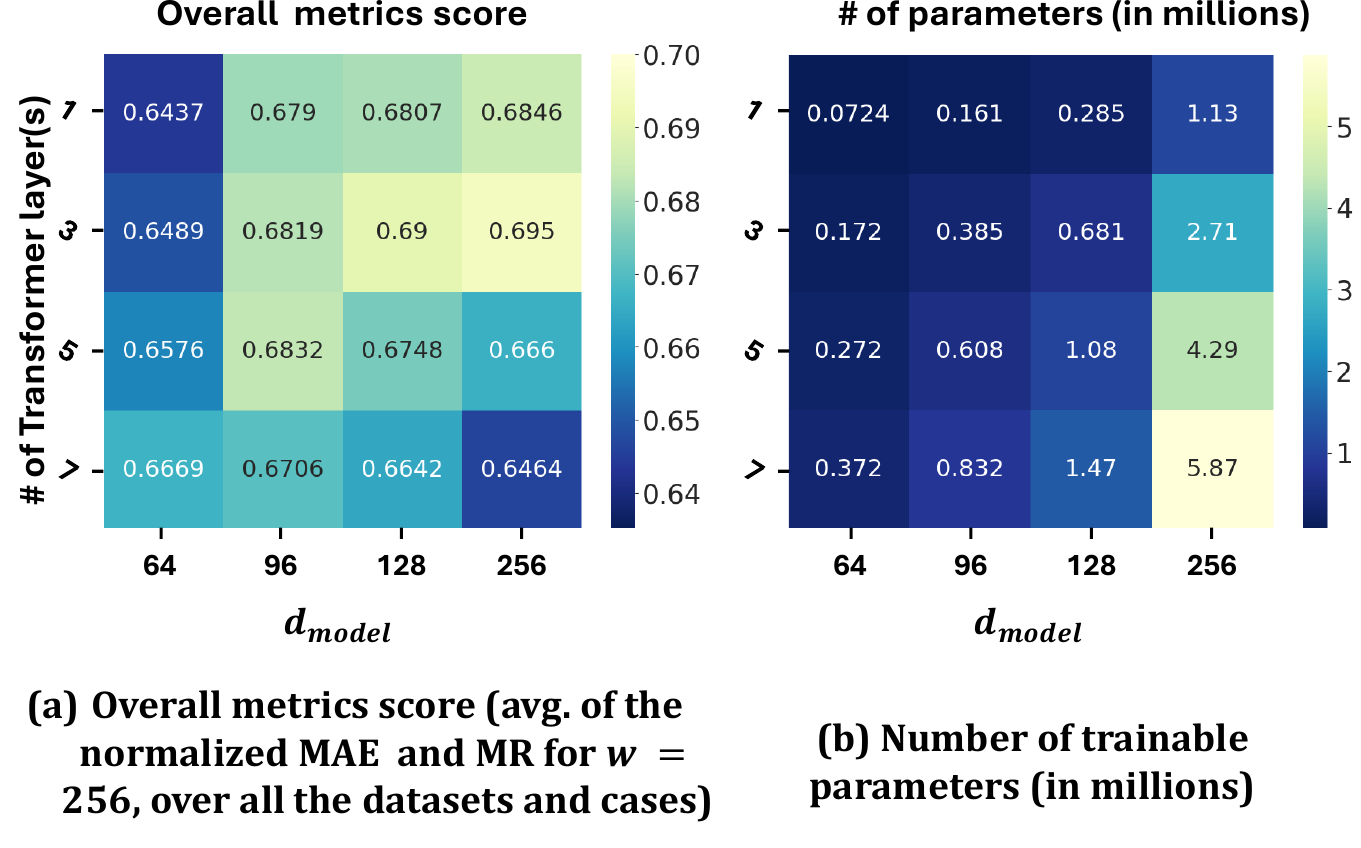}
    \caption{Influence of the number of Transformer layer(s) according to the inner model dimension ($d_{model}$) in NILMFormer on (a) the overall metrics disaggregation score; (b) the number of learnable parameters.}
    \label{fig:hyperparamstudy} 
\end{figure}

We reported the results using the overall metrics score, computed by averaging the 2 metrics (MAE and MR, averaged to the same range) and averaging the scores across the datasets and cases. 
The heatmap in Figure~\ref{fig:hyperparamstudy} (a) indicates that combining 3 layers with an inner dimension of $d=256$ yields the best disaggregation performance. 
However, as reported in Figure~\ref{fig:hyperparamstudy} (b), this combination induces a higher number of trainable parameters (3.55 million).
Therefore, due to the real-world deployment of our solution, we opted for a more efficient configuration: 3 layers with an inner dimension $d=96$.
This combination offers a good balance between accuracy and a reduced number of parameters.

{\footnotesize
\begin{table}[tb]
\caption{NILMFormer hyperparameters}
\label{table:hyperparameters}
\begin{tabular}{c|c|c}
    \toprule
    \multicolumn{3}{c}{\textbf{Hyperparameters}} \\
    \midrule
    \multirow{4}{*}{Embedding block} & $\sharp$ ResBlock & 4 \\
    & $\sharp$ filters & 72 \\
    & kernel size & 3 \\
    & Dilation rate & $\{1, 2, 3, 4\}$ \\
    \hline
    \multirow{6}{*}{Transformer block} & $\sharp$ Transformer Layers & 3 \\
    & d\_model & 96 \\
    & $\sharp$ heads & 8 \\
    & PFFN ratio & 4 \\
    & PFFN activation & GeLU \\
    & Dropout & 0.2 \\
    \hline
    \multirow{2}{*}{Head} & $\sharp$ filters & 128 \\
    & kernel size & 3 \\
    
    \bottomrule
\end{tabular}
\end{table}
}

\section{Models Complexity}

We examine the number of trainable parameters to compare the complexity of the different NILM baselines used in our experiments.
Since this number is influenced by the subsequence window length for certain baselines, we present the number of parameters according to the window length (with $w=\{128, 256, 512\}$.
The results, reported in Table~\ref{table:learnableparams}, indicate that FCN and BiGRU are the smallest models in terms of trainable parameters.
However, despite the use of the Transformer architecture, NILMFormer's number of trainable parameters is kept small compared to the other baselines.
Specifically, the second and third-best baselines, BERT4NILM and STNILM, contain over 1 million and 11 million parameters, respectively.

{\small
\begin{table}[tb]
\caption{Number of learnable parameters (in millions) according to input sequence window length.}
\label{table:learnableparams}
\begin{tabular}{c|c|c|c}
    \toprule
    \multirow{2}{*}{{\bf Models}}  & \multicolumn{3}{c}{\textbf{Input subsequences $w$ length}} \\
    & $w=128$ & $w=256$ & $w=512$ \\
    \midrule
    NILMFormer & 0.385 & 0.385 & 0.385 \\
    BERT4NILM & 1.893 & 1.91 & 1.943 \\
    STNILM & 11.359 & 11.375 & 11.408 \\
    BiGRU & 0.423 & 0.423 & 0.423 \\
    Energformer & 0.567 & 0.567 & 0.567 \\
    FCN & 0.169 & 0.3 & 0.562 \\
    UNet-NILM & 2.414 & 2.676 & 3.202 \\
    TSILNet & 17.402 & 34.245 & 67.931 \\
    BiLSTM & 4.517 & 8.728 & 17.15 \\
    DiffNILM & 9.221 & 9.221 & 9.221 \\
    DAResNet & 0.331 & 0.462 & 0.724 \\
    \bottomrule
\end{tabular}
\end{table}
}

{\footnotesize
\begin{table}[tb]
\caption{Details of training hyperparameters}
\label{table:trainingparam}
\begin{tabular}{c|c}
    \toprule
    \multicolumn{2}{c}{Hyperparameters} \\
    \midrule
    Init. learning rate & 1e-4 \\
    Scheduler & ReduceLROnPlateau \\
    Patience reduce lr & 5 \\
    Batch size & 64 \\
    Max. number of epochs & 50 \\
    Early stopping epochs & 10 \\
    \bottomrule
\end{tabular}
\end{table}
}

\section{Reproducibility}
\label{appendix:reproducibility}

All reported scores are averaged over three household-disjoint train/test splits generated with seeds $\{0,1,2\}$.

\noindent\textbf{[Deep-learning Baselines]}  
All DL baselines were re-implemented in PyTorch 2.5.1, except \textsc{BERT4NILM}~\cite{bert4nilm_2020}, for which we used the authors’ code.  
Training hyper-parameters appear in Table~\ref{table:trainingparam}.  
Models were optimised with Adam, a decaying learning-rate scheduler, and early stopping.

\noindent\textbf{[Other Baselines]}  
ROCKET leverages the Aeon implementation~\cite{aeon}, while XGBoost uses the official Python package~\cite{dmlcxgboost}.

\end{document}

%% file: resultstablesmall.tex
\begin{table*}
\caption{\label{table:resultssmall}{Overall results for the different disaggregation cases, datasets, and subsequences window length. Each reported result is the average score obtained for 3 runs.} The best score is shown in \textbf{\textcolor{blue}{bold}}, and the second best is \underline{\textcolor{purple}{underlined}}.}
\begin{adjustbox}{width=\textwidth,center}
\centering
{\scriptsize
\begin{tabular}{ccc||cc|cc|cc|cc|cc|cc|cc|cc|cc|cc|cc} 
\toprule
     \multicolumn{3}{c||}{Model} & \multicolumn{2}{c|}{NILMFormer} & \multicolumn{2}{c|}{BERT4NILM} & \multicolumn{2}{c|}{STNILM} & \multicolumn{2}{c|}{BiGRU} & \multicolumn{2}{c|}{Energformer} & \multicolumn{2}{c|}{FCN} & \multicolumn{2}{c|}{UNet NILM} & \multicolumn{2}{c|}{TSILNet} & \multicolumn{2}{c|}{BiLSTM} & \multicolumn{2}{c|}{DiffNILM} & \multicolumn{2}{c}{DAResNet} \\
     \multicolumn{2}{c}{Dataset and case} & Win & MAE & MR & MAE & MR & MAE & MR & MAE & MR & MAE & MR & MAE & MR & MAE & MR & MAE & MR & MAE & MR & MAE & MR & MAE & MR \\

\midrule

\multirow{15}{*}{\rotatebox{90}{\scriptsize{UKDALE}}}  & 
\multirow{3}{*}{\rotatebox{0}{\scriptsize{Dishwasher}}} 
    & 128 & \textbf{\textcolor{blue}{22.9}} & \textbf{\textcolor{blue}{0.534}} & 32.2 & 0.27 & 33.3 & 0.257 & \underline{\textcolor{purple}{30.8}} & \underline{\textcolor{purple}{0.326}} & 41.4 & 0.096 & 39.8 & 0.171 & 36.4 & 0.28 & 38.7 & 0.267 & 36.0 & 0.319 & 59.2 & 0.018 & 48.0 & 0.293 \\
    & & 256 & \textbf{\textcolor{blue}{16.7}} & \textbf{\textcolor{blue}{0.626}} & 34.4 & 0.239 & 31.6 & 0.287 & \underline{\textcolor{purple}{30.9}} & \underline{\textcolor{purple}{0.359}} & 40.4 & 0.099 & 43.2 & 0.145 & 40.6 & 0.268 & 48.6 & 0.203 & 44.3 & 0.273 & 87.6 & 0.022 & 66.0 & 0.209 \\
    & & 512 & \textbf{\textcolor{blue}{24.1}} & \textbf{\textcolor{blue}{0.442}} & 34.4 & 0.265 & 31.2 & 0.314 & \underline{\textcolor{purple}{27.2}} & \underline{\textcolor{purple}{0.408}} & 45.2 & 0.097 & 45.3 & 0.171 & 47.8 & 0.205 & 53.4 & 0.192 & 57.4 & 0.255 & 95.3 & 0.036 & 126.5 & 0.085 \\

\cline{2-25}
& \multirow{3}{*}{\scriptsize{Fridge}}
    & 128 & 31.6 & 0.364 & 25.6 & 0.473 & 36.9 & 0.364 & 36.7 & 0.363 & 26.5 & 0.471 & 25.8 & 0.493 & \textbf{\textcolor{blue}{23.3}} & \textbf{\textcolor{blue}{0.536}} & \underline{\textcolor{purple}{24.6}} & 0.528 & \underline{\textcolor{purple}{24.6}} & \underline{\textcolor{purple}{0.533}} & 28.8 & 0.494 & 92.9 & 0.266 \\
     & & 256 & 34.4 & 0.291 & \underline{\textcolor{purple}{27.0}} & 0.447 & 36.9 & 0.364 & 36.8 & 0.365 & 32.3 & 0.384 & 27.8 & \underline{\textcolor{purple}{0.458}} & \textbf{\textcolor{blue}{26.0}} & \textbf{\textcolor{blue}{0.485}} & 33.3 & 0.412 & 36.9 & 0.366 & 72.6 & 0.256 & 33.8 & 0.444 \\
     & & 512 & 33.3 & 0.315 & \textbf{\textcolor{blue}{26.6}} & \underline{\textcolor{purple}{0.452}} & 36.9 & 0.364 & 36.7 & 0.364 & 33.3 & 0.427 & \underline{\textcolor{purple}{28.0}} & \textbf{\textcolor{blue}{0.478}} & 29.8 & 0.425 & 37.0 & 0.365 & 37.0 & 0.366 & 83.9 & 0.225 & 122.0 & 0.159 \\

\cline{2-25}
& \multirow{3}{*}{\rotatebox{0}{\scriptsize{Kettle}}}  
    & 128 & \textbf{\textcolor{blue}{8.7}} & \textbf{\textcolor{blue}{0.705}} & 11.0 & 0.642 & \underline{\textcolor{purple}{9.8}} & 0.665 & \underline{\textcolor{purple}{9.8}} & \underline{\textcolor{purple}{0.67}} & 13.5 & 0.595 & 17.4 & 0.515 & 16.2 & 0.535 & 18.9 & 0.496 & 19.0 & 0.512 & 15.3 & 0.573 & 94.4 & 0.265 \\
    & & 256 & 11.0 & 0.635 & 12.6 & 0.608 & \textbf{\textcolor{blue}{9.8}} & \textbf{\textcolor{blue}{0.664}} & \underline{\textcolor{purple}{10.0}} & \textbf{\textcolor{blue}{0.664}} & 10.1 & 0.663 & 22.0 & 0.438 & 25.8 & 0.406 & 23.9 & 0.437 & 32.4 & 0.364 & 63.1 & 0.046 & 38.8 & 0.278 \\
    & & 512 & \textbf{\textcolor{blue}{9.8}} & \textbf{\textcolor{blue}{0.666}} & 13.6 & 0.583 & \underline{\textcolor{purple}{10.8}} & \underline{\textcolor{purple}{0.632}} & 11.6 & 0.618 & 12.3 & 0.602 & 29.2 & 0.353 & 43.7 & 0.223 & 34.6 & 0.314 & 60.4 & 0.126 & 88.3 & 0.037 & 113.8 & 0.068 \\

\cline{2-25}
& \multirow{3}{*}{\scriptsize{Microwave}}
    & 128 & 10.1 & 0.144 & \textbf{\textcolor{blue}{8.7}} & \textbf{\textcolor{blue}{0.22}} & \underline{\textcolor{purple}{8.9}} & 0.185 & 12.8 & 0.088 & 9.0 & \underline{\textcolor{purple}{0.201}} & 14.6 & 0.116 & 13.8 & 0.062 & 15.9 & 0.051 & 15.6 & 0.045 & 16.7 & 0.011 & 23.5 & 0.044 \\
     & & 256 & \underline{\textcolor{purple}{10.1}} & 0.13 & \textbf{\textcolor{blue}{9.1}} & \textbf{\textcolor{blue}{0.17}} & 11.2 & 0.099 & 14.3 & 0.043 & 11.0 & \underline{\textcolor{purple}{0.149}} & 15.2 & 0.075 & 14.3 & 0.04 & 16.3 & 0.032 & 16.1 & 0.028 & 37.3 & 0.01 & 29.1 & 0.036 \\
     & & 512 & \textbf{\textcolor{blue}{8.3}} & \textbf{\textcolor{blue}{0.187}} & \underline{\textcolor{purple}{10.4}} & 0.111 & 10.9 & 0.066 & 12.4 & 0.04 & 10.5 & \underline{\textcolor{purple}{0.129}} & 17.0 & 0.027 & 14.7 & 0.018 & 16.0 & 0.022 & 15.9 & 0.012 & 56.0 & 0.009 & 80.6 & 0.013 \\

\cline{2-25}
& \multirow{3}{*}{\rotatebox{0}{\scriptsize{Washer}}} 
    & 128 & \textbf{\textcolor{blue}{12.4}} & \textbf{\textcolor{blue}{0.247}} & \underline{\textcolor{purple}{16.4}} & \underline{\textcolor{purple}{0.137}} & 21.4 & 0.114 & 20.6 & 0.109 & 29.4 & 0.136 & 24.9 & 0.114 & 22.7 & 0.109 & 28.5 & 0.06 & 28.6 & 0.068 & 34.8 & 0.021 & 35.9 & 0.075 \\
    & & 256 & \textbf{\textcolor{blue}{8.5}} & \textbf{\textcolor{blue}{0.358}} & 26.3 & 0.084 & 26.6 & 0.081 & \underline{\textcolor{purple}{19.0}} & 0.093 & 32.0 & \underline{\textcolor{purple}{0.116}} & 32.9 & 0.089 & 25.4 & 0.093 & 34.5 & 0.051 & 32.9 & 0.055 & 50.1 & 0.014 & 49.2 & 0.042 \\
    & & 512 & \textbf{\textcolor{blue}{10.8}} & \textbf{\textcolor{blue}{0.268}} & 31.5 & 0.09 & 26.5 & \underline{\textcolor{purple}{0.098}} & \underline{\textcolor{purple}{21.1}} & 0.086 & 31.4 & 0.086 & 38.3 & 0.058 & 29.2 & 0.066 & 34.8 & 0.052 & 39.3 & 0.044 & 66.1 & 0.013 & 90.5 & 0.019 \\

\midrule

\multirow{12}{*}{\rotatebox{90}{\scriptsize{REFIT}}}  
& \multirow{3}{*}{\rotatebox{0}{\scriptsize{Dishwasher}}} 
    & 128 & \textbf{\textcolor{blue}{29.1}} & \textbf{\textcolor{blue}{0.332}} & 49.3 & 0.081 & 41.1 & 0.136 & 44.5 & 0.169 & \underline{\textcolor{purple}{35.8}} & 0.082 & 39.3 & \underline{\textcolor{purple}{0.218}} & 49.6 & 0.166 & 47.0 & 0.134 & 70.0 & 0.091 & 69.0 & 0.018 & 242.2 & 0.044 \\
    & & 256 & 45.5 & \textbf{\textcolor{blue}{0.312}} & \underline{\textcolor{purple}{44.2}} & 0.145 & \textbf{\textcolor{blue}{40.5}} & 0.149 & 70.4 & 0.069 & 54.9 & 0.084 & 50.2 & \underline{\textcolor{purple}{0.151}} & 59.2 & 0.093 & 69.4 & 0.065 & 72.7 & 0.057 & 74.3 & 0.02 & 125.7 & 0.049 \\
    & & 512 & \textbf{\textcolor{blue}{42.7}} & \textbf{\textcolor{blue}{0.205}} & 56.0 & 0.117 & \underline{\textcolor{purple}{46.3}} & 0.121 & 75.9 & 0.072 & 64.6 & 0.112 & 52.8 & \underline{\textcolor{purple}{0.134}} & 59.1 & 0.086 & 72.6 & 0.046 & 90.2 & 0.037 & 83.1 & 0.022 & 231.9 & 0.042 \\

\cline{2-25}
& \multirow{3}{*}{\rotatebox{0}{\scriptsize{Kettle}}}  
    & 128 & \underline{\textcolor{purple}{9.3}} & \underline{\textcolor{purple}{0.522}} & 16.9 & 0.437 & \textbf{\textcolor{blue}{8.8}} & \textbf{\textcolor{blue}{0.529}} & 9.4 & 0.508 & 12.4 & 0.426 & 15.5 & 0.387 & 15.7 & 0.368 & 17.6 & 0.348 & 18.0 & 0.35 & 33.8 & 0.009 & 53.4 & 0.121 \\
    & & 256 & 12.5 & 0.454 & \textbf{\textcolor{blue}{9.2}} & \textbf{\textcolor{blue}{0.504}} & \underline{\textcolor{purple}{9.8}} & \underline{\textcolor{purple}{0.503}} & 10.0 & 0.496 & 16.0 & 0.359 & 18.2 & 0.332 & 19.1 & 0.31 & 18.9 & 0.312 & 25.4 & 0.249 & 34.1 & 0.141 & 54.9 & 0.123 \\
    & & 512 & \textbf{\textcolor{blue}{8.6}} & \textbf{\textcolor{blue}{0.526}} & \underline{\textcolor{purple}{10.6}} & \underline{\textcolor{purple}{0.462}} & 11.0 & 0.439 & 19.3 & 0.329 & 17.7 & 0.306 & 22.5 & 0.259 & 28.3 & 0.205 & 26.2 & 0.233 & 42.5 & 0.054 & 43.2 & 0.014 & 119.9 & 0.051 \\

\cline{2-25}
& \multirow{3}{*}{\scriptsize{Microwave}}
    & 128 & \textbf{\textcolor{blue}{5.7}} & \textbf{\textcolor{blue}{0.11}} & 10.7 & \underline{\textcolor{purple}{0.074}} & 11.9 & 0.04 & \underline{\textcolor{purple}{9.1}} & 0.04 & 10.0 & 0.046 & 12.9 & 0.05 & 12.9 & 0.043 & 10.1 & 0.022 & 10.2 & 0.023 & 22.2 & 0.006 & 43.4 & 0.014 \\
    & & 256 & \textbf{\textcolor{blue}{9.7}} & \underline{\textcolor{purple}{0.05}} & 10.7 & \textbf{\textcolor{blue}{0.056}} & \textbf{\textcolor{blue}{9.7}} & 0.029 & 10.0 & 0.023 & 10.2 & 0.032 & 10.3 & 0.028 & 13.8 & 0.023 & 9.8 & 0.019 & 11.1 & 0.017 & 21.7 & 0.007 & 77.6 & 0.009 \\
    & & 512 & \textbf{\textcolor{blue}{6.4}} & \textbf{\textcolor{blue}{0.082}} & 10.7 & \underline{\textcolor{purple}{0.029}} & \underline{\textcolor{purple}{8.0}} & 0.024 & 9.7 & 0.015 & 9.9 & 0.027 & 11.4 & 0.015 & 11.3 & 0.013 & 9.5 & 0.014 & 9.2 & 0.007 & 31.0 & 0.006 & 540.2 & 0.007 \\

\cline{2-25}
& \multirow{3}{*}{\scriptsize{Washer}}
    & 128 & \textbf{\textcolor{blue}{18.9}} & \textbf{\textcolor{blue}{0.25}} & 33.6 & \underline{\textcolor{purple}{0.172}} & 33.0 & 0.126 & \underline{\textcolor{purple}{30.6}} & 0.115 & 36.9 & 0.115 & 35.5 & 0.112 & 42.7 & 0.1 & 35.8 & 0.109 & 40.9 & 0.097 & 48.4 & 0.023 & 101.4 & 0.048 \\
    & & 256 & \textbf{\textcolor{blue}{22.3}} & \textbf{\textcolor{blue}{0.172}} & 34.8 & 0.109 & 30.9 & \underline{\textcolor{purple}{0.125}} & 34.0 & 0.1 & \underline{\textcolor{purple}{30.7}} & 0.109 & 42.0 & 0.098 & 42.4 & 0.07 & 42.8 & 0.077 & 40.0 & 0.076 & 57.7 & 0.026 & 166.5 & 0.029 \\
    & & 512 & \textbf{\textcolor{blue}{20.6}} & \textbf{\textcolor{blue}{0.168}} & 31.0 & 0.09 & \underline{\textcolor{purple}{29.0}} & \underline{\textcolor{purple}{0.11}} & 37.4 & 0.075 & 32.8 & 0.097 & 40.1 & 0.094 & 35.7 & 0.086 & 40.5 & 0.069 & 43.0 & 0.063 & 50.3 & 0.027 & 188.4 & 0.029 \\

\midrule

\multirow{12}{*}{\rotatebox{90}{\scriptsize{EDF1}}} 
    & \multirow{3}{*}{\rotatebox{0}{\scriptsize{Heater}}}  
    & 128 & \textbf{\textcolor{blue}{283.4}} & \textbf{\textcolor{blue}{0.52}} & 312.1 & \underline{\textcolor{purple}{0.457}} & \underline{\textcolor{purple}{310.3}} & 0.44 & 342.7 & 0.435 & 354.0 & 0.445 & 328.0 & 0.428 & 352.8 & 0.437 & 374.6 & 0.416 & 345.0 & 0.425 & 351.6 & 0.434 & 529.7 & 0.305 \\
    & & 256 & \textbf{\textcolor{blue}{263.6}} & \textbf{\textcolor{blue}{0.525}} & 297.2 & \underline{\textcolor{purple}{0.458}} & \underline{\textcolor{purple}{290.8}} & 0.456 & 323.0 & 0.425 & 371.9 & 0.414 & 316.6 & 0.437 & 359.0 & 0.402 & 337.7 & 0.405 & 330.5 & 0.415 & 333.5 & 0.442 & 533.6 & 0.296 \\
    & & 512 & \textbf{\textcolor{blue}{255.8}} & \textbf{\textcolor{blue}{0.521}} & 283.3 & \underline{\textcolor{purple}{0.456}} & \underline{\textcolor{purple}{275.3}} & \underline{\textcolor{purple}{0.456}} & 321.1 & 0.411 & 368.7 & 0.386 & 307.9 & 0.424 & 323.1 & 0.403 & 314.6 & 0.396 & 330.0 & 0.395 & 389.2 & 0.336 & 497.2 & 0.26 \\

\cline{2-25}
& \multirow{3}{*}{\rotatebox{0}{\scriptsize{Heatpump}}}  
    & 128 & \textbf{\textcolor{blue}{270.5}} & \textbf{\textcolor{blue}{0.548}} & 288.5 & 0.504 & 283.3 & 0.523 & 282.0 & 0.526 & \underline{\textcolor{purple}{274.9}} & \underline{\textcolor{purple}{0.53}} & 328.1 & 0.475 & 362.9 & 0.441 & 326.6 & 0.462 & 337.4 & 0.446 & 344.0 & 0.452 & 507.4 & 0.356 \\
    & & 256 & \textbf{\textcolor{blue}{276.2}} & \textbf{\textcolor{blue}{0.545}} & 292.0 & 0.504 & 282.2 & 0.501 & 288.1 & 0.507 & \underline{\textcolor{purple}{280.4}} & \underline{\textcolor{purple}{0.51}} & 353.1 & 0.43 & 364.0 & 0.422 & 345.6 & 0.451 & 370.9 & 0.412 & 304.9 & 0.486 & 542.6 & 0.272 \\
    & & 512 & \textbf{\textcolor{blue}{258.1}} & \textbf{\textcolor{blue}{0.533}} & 295.9 & 0.475 & \underline{\textcolor{purple}{290.3}} & \underline{\textcolor{purple}{0.489}} & 303.0 & 0.457 & 299.2 & 0.48 & 392.7 & 0.374 & 349.9 & 0.394 & 363.3 & 0.388 & 377.5 & 0.375 & 362.9 & 0.397 & 579.7 & 0.272 \\

\cline{2-25}
& \multirow{3}{*}{\rotatebox{0}{\scriptsize{WaterHeater}}}  
    & 128 & \textbf{\textcolor{blue}{88.2}} & \textbf{\textcolor{blue}{0.686}} & 113.1 & 0.613 & 134.9 & 0.566 & \underline{\textcolor{purple}{112.5}} & \underline{\textcolor{purple}{0.619}} & 135.6 & 0.551 & 185.2 & 0.46 & 224.9 & 0.364 & 169.8 & 0.494 & 178.9 & 0.478 & 131.0 & 0.55 & 268.8 & 0.361 \\
    & & 256 & \textbf{\textcolor{blue}{91.8}} & \textbf{\textcolor{blue}{0.676}} & 126.3 & 0.591 & 136.4 & 0.564 & 114.2 & 0.614 & 146.5 & 0.521 & 207.6 & 0.428 & 238.9 & 0.337 & 203.6 & 0.443 & 223.7 & 0.4 & \underline{\textcolor{purple}{104.8}} & \underline{\textcolor{purple}{0.639}} & 342.2 & 0.255 \\
    & & 512 & \textbf{\textcolor{blue}{92.4}} & \textbf{\textcolor{blue}{0.668}} & \underline{\textcolor{purple}{111.2}} & \underline{\textcolor{purple}{0.615}} & 138.1 & 0.55 & 118.5 & 0.601 & 181.0 & 0.464 & 223.7 & 0.392 & 237.9 & 0.34 & 236.9 & 0.372 & 274.8 & 0.319 & 126.7 & 0.587 & 356.0 & 0.256 \\

\cline{2-25}
& \multirow{3}{*}{\rotatebox{0}{\scriptsize{White Appl.}}} 
    & 128 & \textbf{\textcolor{blue}{90.8}} & \textbf{\textcolor{blue}{0.209}} & 105.2 & 0.203 & 111.6 & 0.171 & \underline{\textcolor{purple}{99.3}} & \textbf{\textcolor{blue}{0.209}} & 109.8 & 0.179 & 120.9 & 0.158 & 116.9 & 0.139 & 127.5 & 0.123 & 128.3 & 0.145 & 134.1 & 0.083 & 164.5 & 0.105 \\
    & & 256 & \textbf{\textcolor{blue}{86.1}} & \textbf{\textcolor{blue}{0.231}} & 104.2 & 0.189 & \underline{\textcolor{purple}{99.3}} & \underline{\textcolor{purple}{0.191}} & 106.0 & 0.176 & 114.8 & 0.133 & 130.4 & 0.143 & 119.8 & 0.144 & 131.3 & 0.107 & 133.1 & 0.119 & 126.5 & 0.069 & 206.8 & 0.098 \\
    & & 512 & \textbf{\textcolor{blue}{79.5}} & \textbf{\textcolor{blue}{0.215}} & 94.2 & \underline{\textcolor{purple}{0.178}} & \underline{\textcolor{purple}{89.7}} & \underline{\textcolor{purple}{0.178}} & 102.8 & 0.128 & 107.9 & 0.123 & 114.8 & 0.103 & 115.2 & 0.099 & 116.8 & 0.102 & 124.0 & 0.088 & 203.2 & 0.06 & 192.2 & 0.071 \\

\midrule

\multirow{3}{*}{\rotatebox{90}{\scriptsize{EDF2}}} 
& \multirow{3}{*}{\rotatebox{0}{\scriptsize{EV}}} 
    & 128 & \textbf{\textcolor{blue}{102.5}} & \textbf{\textcolor{blue}{0.617}} & 169.2 & 0.472 & \underline{\textcolor{purple}{109.9}} & \underline{\textcolor{purple}{0.57}} & 210.4 & 0.402 & 134.8 & 0.507 & 189.5 & 0.428 & 245.7 & 0.384 & 242.5 & 0.372 & 238.4 & 0.371 & 286.3 & 0.322 & 434.5 & 0.233 \\
    & & 256 & \textbf{\textcolor{blue}{111.8}} & \textbf{\textcolor{blue}{0.582}} & 168.6 & 0.456 & \underline{\textcolor{purple}{119.1}} & \underline{\textcolor{purple}{0.553}} & 227.8 & 0.36 & 180.1 & 0.396 & 232.2 & 0.381 & 262.5 & 0.31 & 299.4 & 0.303 & 301.2 & 0.313 & 213.8 & 0.404 & 932.8 & 0.097 \\
    & & 512 & \textbf{\textcolor{blue}{114.3}} & \textbf{\textcolor{blue}{0.616}} & 152.5 & 0.544 & \underline{\textcolor{purple}{130.6}} & \underline{\textcolor{purple}{0.597}} & 327.1 & 0.304 & 147.3 & 0.556 & 323.2 & 0.326 & 398.4 & 0.205 & 357.7 & 0.206 & 481.4 & 0.122 & 295.3 & 0.354 & 1897.1 & 0.046 \\

\midrule
\multicolumn{3}{c||}{Avg. Score Per Metrics} & 
\textbf{\textcolor{blue}{70.7}} & \textbf{\textcolor{blue}{0.4}} & 84.5 & \underline{\textcolor{purple}{0.328}} & \underline{\textcolor{purple}{81.2}} & 0.3263 & 94.5 & 0.3042 & 93.2 & 0.2914 & 107.8 & 0.2611 & 116.0 & 0.2412 & 114.5 & 0.2373 & 122.5 & 0.2216 & 122.2 & 0.1838 & 261.3 & 0.1462 \\

\multicolumn{3}{c||}{Avg. Rank Per Metrics} & \textbf{\textcolor{blue}{1.762}} & \textbf{\textcolor{blue}{1.905}} & 3.667 & \underline{\textcolor{purple}{3.476}} & \underline{\textcolor{purple}{3.31}} & 3.762 & 4.262 & 4.667 & 4.881 & 4.714 & 6.286 & 5.714 & 6.857 & 6.81 & 7.262 & 7.69 & 8.143 & 8.143 & 8.833 & 9.238 & 10.738 & 9.881 \\

\multicolumn{3}{c||}{Avg. Rank} &  \multicolumn{2}{c|}{\textbf{\textcolor{blue}{1.833}}} & \multicolumn{2}{c|}{\underline{\textcolor{purple}{3.536}}} & \multicolumn{2}{c|}{3.571} & \multicolumn{2}{c|}{4.464} & \multicolumn{2}{c|}{4.798} & \multicolumn{2}{c|}{6.0} & \multicolumn{2}{c|}{6.833} & \multicolumn{2}{c|}{7.476} & \multicolumn{2}{c|}{8.143} & \multicolumn{2}{c|}{9.036} & \multicolumn{2}{c}{10.31} \\

\bottomrule
\end{tabular}
}
\end{adjustbox}
\end{table*}